\documentclass[10pt,twocolumn,letterpaper]{article}

\usepackage{kotex}
\usepackage[pagenumbers]{iccv} 

\usepackage[dvipsnames]{xcolor}

\usepackage{multirow} 
\usepackage{amsmath} 
\usepackage{bbm} 
\usepackage{siunitx} 
\usepackage{kotex} 
\usepackage{array,graphicx}
\usepackage{float}
\usepackage{amsmath}
\usepackage{mathtools}
\usepackage{bbm}
\usepackage{hhline}
\usepackage{boldline}
\usepackage{tabularx}
\usepackage{float}
\usepackage{comment}
\usepackage{color}
\usepackage{amssymb}
\usepackage{booktabs}
\usepackage{multicol}
\usepackage{microtype}      
\usepackage{xcolor}         
\usepackage{algorithm}
\usepackage{algpseudocode}
\usepackage{pifont}
\usepackage{soul} 
\usepackage{subcaption}
\usepackage{wrapfig}
\usepackage{lipsum}
\usepackage{booktabs}
\usepackage{colortbl}
\usepackage{nicematrix}

\definecolor{iccvblue}{rgb}{0.21,0.49,0.74}
\usepackage[pagebackref,breaklinks,colorlinks,allcolors=iccvblue]{hyperref}

\def\paperID{5259} 
\def\confName{ICCV}
\def\confYear{2025}

\title{Prototypes are Balanced Units for \\ Efficient and Effective Partially Relevant Video Retrieval}

\author{
    WonJun Moon\thanks{This work was done during an internship at NAVER Cloud.} \;\;\; Cheol-Ho Cho \;\;\; Woojin Jun \\
    Sungkyunkwan University\\
    \and  
    Minho Shim \;\;\; Taeoh Kim \;\;\; Inwoong Lee \;\;\; Dongyoon Wee\\
    NAVER Cloud\\
    \and
    Jae-Pil Heo\thanks{Corresponding Author}\\
    Sungkyunkwan University\\
}

\begin{document}
\maketitle
\vspace{-1.cm}
\begin{abstract}
In a retrieval system, simultaneously achieving search accuracy and efficiency is inherently challenging. 
This challenge is particularly pronounced in partially relevant video retrieval~(PRVR), where incorporating more diverse context representations at varying temporal scales for each video enhances accuracy but increases computational and memory costs.
To address this dichotomy, we propose a prototypical PRVR framework that encodes diverse contexts within a video into a fixed number of prototypes.
We then introduce several strategies to enhance text association and video understanding within the prototypes, along with an orthogonal objective to ensure that the prototypes capture a diverse range of content. 
To keep the prototypes searchable via text queries while accurately encoding video contexts, we implement cross- and uni-modal reconstruction tasks. 
The cross-modal reconstruction task aligns the prototypes with textual features within a shared space, while the uni-modal reconstruction task preserves all video contexts during encoding.
Additionally, we employ a video mixing technique to provide weak guidance to further align prototypes and associated textual representations.
Extensive evaluations on TVR, ActivityNet-Captions, and QVHighlights validate the effectiveness of our approach without sacrificing efficiency.
\end{abstract}
\section{Introduction}
\label{sec:intro}

\begin{figure}[t!]
    \centering
    \vspace{-0.2cm}
    \includegraphics[width=0.47\textwidth]{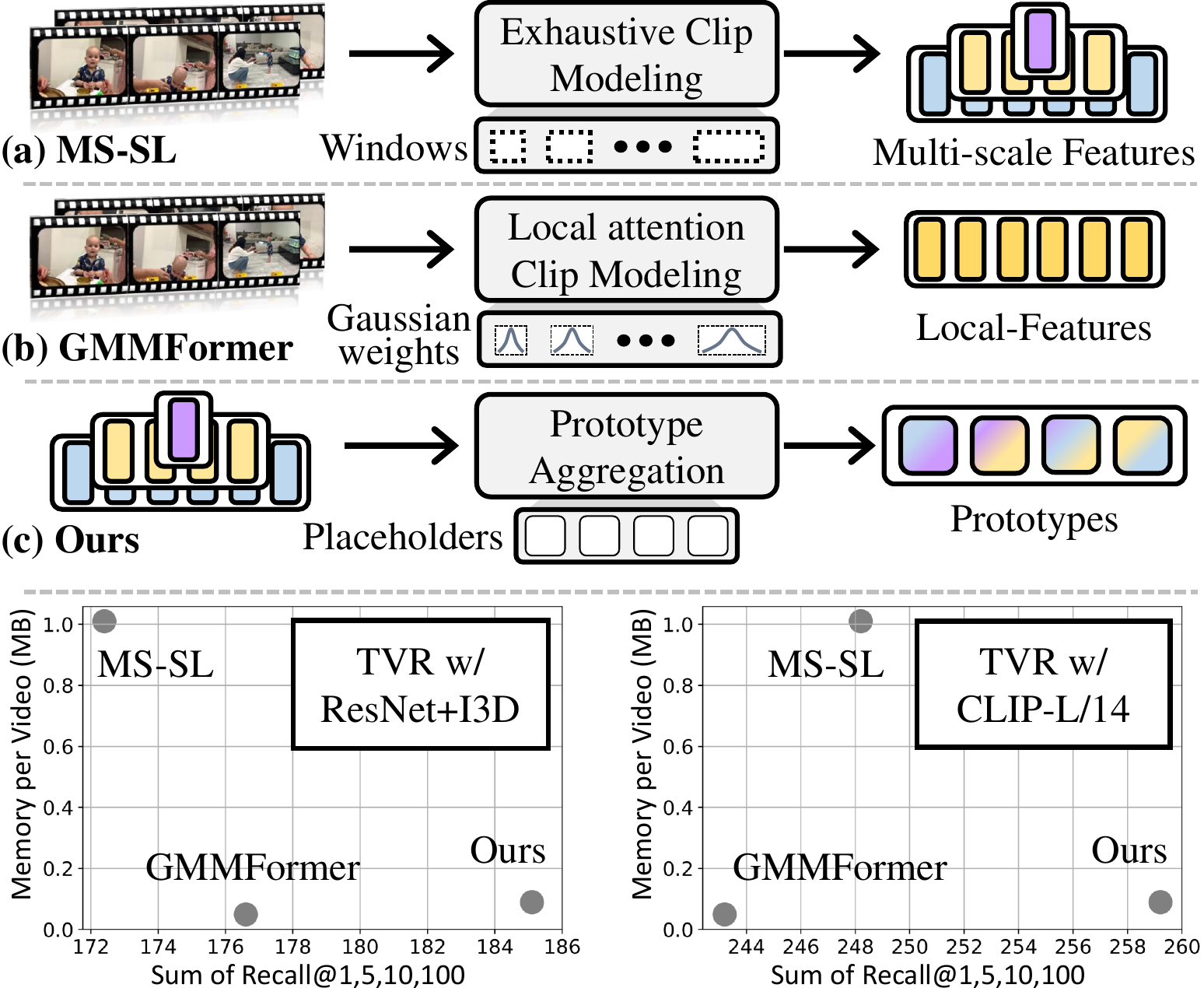}
    \vspace{-0.2cm}
    \caption{
        Comparisons in video encoding process. (a) MS-SL utilizes exhaustive clip modeling based on varying lengths of clip windows to encode contexts at diverse temporal scales. (b) GMMFormer performs a similarity-aware feature aggregation via self-attention constrained with predefined Gaussian kernels to reflect locality. Such an adaptive scheme provides high efficiency.
        (c) To exploit the semantic richness in exhaustive clip modeling without sacrificing efficiency, we learn a fixed number of prototypes that aggregate diverse~(varying lengths or potentially disjointed) contexts within a video.
        (Down) Our proposed method has superiority over previous works in terms of the accuracy-efficiency trade-off.
    }
    \label{fig:motiv}
    \vspace{-0.5cm}
\end{figure}

The sheer volume of video content makes it increasingly challenging for users to find the video they are looking for.
This underscores the need for accurate and efficient video retrieval systems~\cite{clip4clip} that can search through videos consisting of short moments the users are seeking. 
In response to this demand, Partially Relevant Video Retrieval~(PRVR)~\cite{prvr} has been introduced to extend text-video retrieval with long complex videos where only specific segments correspond to the text query.
Typical approaches such as MS-SL~\cite{prvr} prioritize the accuracy by storing lengthy frame-wise~\cite{dldkd} features and exhaustively modeling clips of varying lengths using different window sizes~\cite{pean}, as shown in Fig.~\ref{fig:motiv}~(a).
With these representations, a video that yields the highest similarity to the input text description is retrieved.
While such approaches have made significant progress, they often overlook the importance of efficiency in retrieval systems that operate within extensive databases.


To address efficiency, GMMFormer~\cite{gmmformer} introduced a local clip modeling, which applies Gaussian constraints with predefined variances around selected frames, as shown in Fig.~\ref{fig:motiv}(b).
Specifically, these constraints are integrated within self-attention~\cite{vaswani2017attention, dosovitskiy2021an}.
While this approach improves efficiency by primarily encoding local contexts, it struggles to accurately capture moments when matching queries that describe complex and extended events within video sequences.
This limitation arises because local contexts are confined within fixed-length temporal segments.
Likewise, we observe the trade-off between achieving efficiency and attaining a thorough understanding of contents.


To address the challenge of simultaneously achieving efficiency and embracing diverse contexts, we propose a prototypical network for PRVR.
Specifically, we leverage the advantages of the exhaustive clip modeling strategy to encode both the local and global contexts~(Fig.~\ref{fig:motiv}(a)), while our prototypes effectively aggregate these features across diverse temporal scales into a smaller footprint, as shown in Fig.~\ref{fig:motiv}(c).
Then, only a small set of video prototypes, significantly fewer than the video frame or clip representations, are stored for efficient retrieval.


Despite the potential of prototypes, prototypical learning does not inherently guarantee text associativity or the preservation of all contextual details necessary for retrieval due to its implicit learning nature without explicit guidance on how each prototype should encode specific information.
To address this, we propose several enhancements to prototypical learning to strengthen the association with text descriptions and improve video understanding. 
First, we introduce dual reconstruction tasks, consisting of cross-modal~(video-to-text) and uni-modal~(video-to-video) reconstruction tasks. 
The cross-modal mask reconstruction task enhances the prototypes' comprehension of text queries, while the uni-modal reconstruction task preserves the visual content during prototype aggregation.
In addition, to guarantee alignment between text queries and the visual prototypes that attend to text-relevant video segments without explicit moment supervision, we employ a video mixing strategy that concatenates videos at a higher frame rate.
This strategy trains the model to identify and retrieve a prototype that focuses on the clip within the video that corresponds to the given query.
Finally, with these components combined with the orthogonal loss for prototype diversification, our method achieves state-of-the-art results on TVR, ActivityNet Captions, and QVHighlights datasets with both CNN- and Transformer-based backbones.

\section{Related Work}
\label{sec:relatedwork}
\noindent\textbf{Text-to-Video Retrieval~(T2VR)}
aims to retrieve the query-relevant videos~\cite{dong2019dual, de++, VoP_VR, MMCKD_VR, CLIPPING_VR, Cap4Video_VR, wang2024text, tian2024holistic, DiffVR}. 
One notable stream is utilizing multi-scale spatio-temporal similarity~\cite{UniVR, Pidro, X-clip_VR}.
These works consider the similarity between videos and sentences and also the local similarity between the patches and words for precise retrieval. 
Another stream is to account for the uncertainty in retrieval.
By tackling the many-to-many problems in text-video matching due to diversity in expression in both domains, UATVR~\cite{UATVR} employed Gaussian augmentation to address the uncertainty, and PAU~\cite{AleaVR} exploited uncertainty for training and reranking.
Yet, T2VR assumes that the full-video-level text descriptions are available, limiting their applicability in practice.

\noindent\textbf{Moment Retrieval} localizes partial moments in videos that corresponds to the text descriptions~\cite{anne2017localizing, gao2017tall, soldan2021vlg, mr1, mr2, mr3, mr4,liu2021context,zhang2019exploiting,zhang2019man, qddetr, eatr, univtg,unloc, momentdetr, umt}.
Despite the similarity with PRVR in the need for localizing the relevant clips, moment retrieval only assumes the pair-wise searching problem.
Thus, video-text interactive designs are popularly employed~\cite{xiao2024bridging, cgdetr, trdetr}.

\noindent\textbf{Video Corpus Moment Retrieval} requires multiple videos to be considered at once, thus modality-interactive designs are impractical to be adopted straightforwardly~\cite{hou2024improving, zhang2021multi}.
To maintain efficiency and take advantage of interactive designs, a two-stage pipeline is widely adopted where moment retrieval is implemented after the video retrieval~\cite{conquer,vcmr}.
The difference to PRVR is the presence of localization labels.


\begin{figure*}[t!]
    \centering
    \vspace{-0.3cm}
    \includegraphics[width=0.99\textwidth]{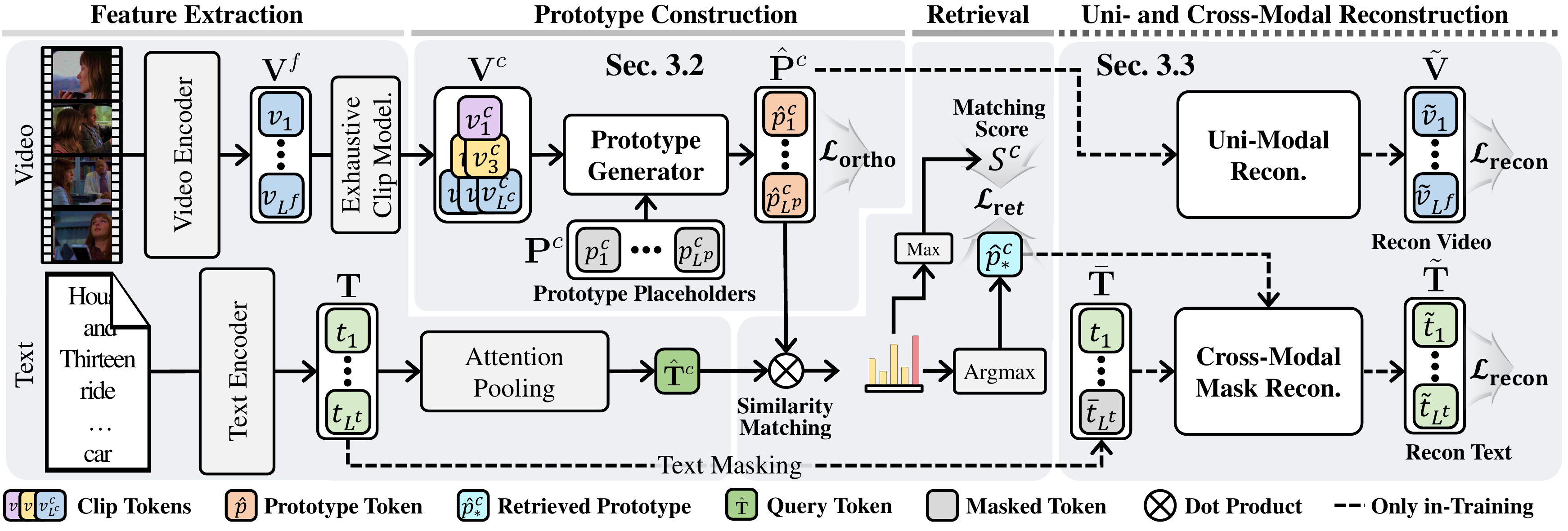}
    \vspace{-0.4cm}
    \caption{An overview of the clip branch in our prototypical framework which is consistent with the frame branch except for the existence of exhaustive clip modeling.
    For the video stream, prototype aggregation is implemented upon clip features formed by an exhaustive clip modeling strategy.
    Text queries are encoded and aggregated through an attention-pooling layer.
    Finally, the similarity matching between the visual prototypes $\hat{\mathbf{P}}^c$ and query token $\hat{\mathbf{T}}^c$ is implemented to calculate the text-to-video score $S^c$ for retrieval.
    On the right side, uni- and cross-modal reconstructions are performed with constructed prototypes during training. 
    Only the retrieved prototype with the maximum similarity is utilized to reconstruct masked text word~(cross-modal) while all prototypes are exploited to reconstruct video frames~(uni-modal).
     }
    \label{fig:main_figure}
    \vspace{-0.2cm}
\end{figure*}

\noindent\textbf{Partially Relevant Video Retrieval}
shares the practical objective of VCMR in retrieving partially relevant videos~\cite{dldkd, pean}. 
Yet, PRVR differs as it assumes that exact moments of relevant clips in the video are not available due to huge costs.
This introduces a more challenging scenario, requiring the implicit identification of text-relevant segments to achieve accurate video-text alignment.
To explore diverse semantics in a video, MS-SL~\cite{prvr} considered encoding all possible partial videos with varying temporal intervals.
However, since such a strategy sacrifices efficiency in terms of both capacity and inference speed, recent works focus on improving efficiency. 
GMMFormer~\cite{gmmformer} aimed to learn the local context in every clip representation by applying Gaussian-distributed coefficients to enforce each clip to attend more to nearby clips in the temporal attention layer.
QASIR~\cite{qasir} also addressed the efficiency with super-images that comprise several frame images.
We share the motivation with these works to enhance efficiency but without sacrificing the capability of encoding complex contexts.


Prototypical learning typically offers two key benefits: efficiency and the ability to organize context-wise representations.
Due to these benefits, numerous variants of prototypical learning have been developed and applied across diverse machine learning domains~\cite{ProST, okazawa2022interclass, wang2019panet, detr, tdm, BLIP2, Flamingo, gemaking}. %
Among them, slot-attention and perceiver are two popular approaches with similar architectural designs.
Slot-attention~\cite{slot} introduces slots as object-centric prototypes in synthetic benchmarks, while perceiver~\cite{jaegle2021perceiver} utilizes compact latent bottlenecks to efficiently encode a wide range of data types.
Our work extends the prototypical learning to video representation learning for PRVR.
In the following section, we elaborate on our architectural design choices to leverage the efficiency of prototypes, with enhancements including dual reconstruction tasks, weak video-text guidance, and prototype diversification.

\section{Prototypical PRVR}
\label{sec:method}
Without loss of generality, video often contains multiple contexts and most of the time, users are only interested in watching part of the video.
Hence, to provide a reliable text-video retrieval system, it is essential to keep every context in the database despite its huge memory cost.
Although recent work partially addressed the efficiency by storing the modeled local clips, we argue that users' needs are not always predetermined in single short contexts.
In this regard, we propose a prototypical framework for PRVR.
We highlight the advantages of the prototypical framework, which enables the utilization of diverse segments from long untrimmed videos to encode both local and complex contexts without additional computational burden at inference.
This is because the learned contexts are ultimately aggregated into a fixed number of prototypes before storage, ensuring an efficient and compact video representation.
Building on these benefits, we introduce a tailored prototypical framework with further enhancements in text comprehensibility and visual understanding capabilities.




\subsection{Overall Architecture}
Following previous works~\cite{prvr,gmmformer} with frame and clip branches, we adopt dual-branch architecture.
Note that the procedure for each branch is identical except for the existence of an exhaustive clip modeling process~\cite{prvr}\footnote{Constructs all possible video clips from single frames to $n$-frame combination, capturing diverse features across all temporal scales.} for the clip branch.
In Fig.~\ref{fig:main_figure}, we depict an overview of the clip branch.
Given a set of videos, frame features denoted as $\mathbf{V}^f = [v_1, ..., v_{L^f}]$ of $L^f$ frames for each video are encoded through a stack of encoding layers and converted into $L^c$ visual clip features $\mathbf{V}^c = [v^c_1, ..., v^c_{L^c}]$ with the exhaustive clip modeling process.
Then, the visual context of each video is aggregated into $L^p$ prototypes.
On the other hand, a set of text queries, each represented as $\mathbf{T} = [t_1, ..., t_{L^t}]$ of $L^t$ words, are projected and pooled to yield a single query token $\hat{\mathbf{T}}^c$.
Finally, the text query $\hat{\mathbf{T}}^c$ is compared against all the prototypes of each video, and the prototype with the highest similarity~($S^c$) is retrieved to represent how each video corresponds to the text query for retrieval training.
Note that we denote the prototype with the maximum similarity as $\hat{p}_*^c$.
For inference, we compute the similarity between the text query and the prototypes of all stored videos.
Then, the video corresponding to the prototype with the highest similarity is retrieved.
For the remainder of the paper, we omit the branch notation for video branch, as they share the same operations.


\subsection{Prototype Construction}
\label{Sec.prototypeconstruction}


Motivated by efficiency and semantic encoding capability of prototypical learning~\cite{jaegle2021perceiver, slot}, our prototype generator operates upon $L^p$ globally shared learnable prototypes, denoted as $\mathbf{P} = [p_1, ..., p_{L^p}]$.
Our prototype generator is streamlined in its design, consisting of only the cross-attention.
The rationale behind solely using cross-attention is that independent prototypes are more effective for retrieval as they tend to capture specific events more precisely.
This simplicity also avoids unnecessary operations, contributing to more efficient retrieval.
With this design, prototypes are gradually adapted to each instance over $K$ iterations, collectively encapsulating the specific contexts within the video.
Formally, $k$-th iteration for aggregating video features $\mathbf{V}$ is expressed as:
\begin{equation}
\hat{\mathbf{P}}^{(k)} = \text{CA}(\hat{\mathbf{P}}^{(k-1)}\;; \mathbf{V}\;; \mathbf{V} ), 
\end{equation}
where CA denotes cross-attention and $\hat{\mathbf{P}}^0$ is initialized with $\mathbf{P}$.
For the rest of the paper, we abbreviate $\hat{\mathbf{P}}^{(K)}$ as $\hat{\mathbf{P}}$.

As discussed, the key benefit of a prototypical framework lies in its efficient management of diverse visual content across various temporal scales.
However, the implicit encoding process may fail to construct text-associative prototypes or fully preserve all visual details.
To overcome these limitations, we propose several strategies to refine the prototypes.


\subsection{Dual Reconstruction with Prototypes}
\label{Sec.reconstruction}
Reconstruction is a widely employed technique to mitigate information loss during the prototype generation process~\cite{slot}. 
This is particularly important in PRVR where the prototypes must preserve diverse contexts of the video to handle partially relevant text descriptions.

\begin{figure}[t!]
    \centering
    \vspace{-0.2cm}
    \includegraphics[width=0.41\textwidth]{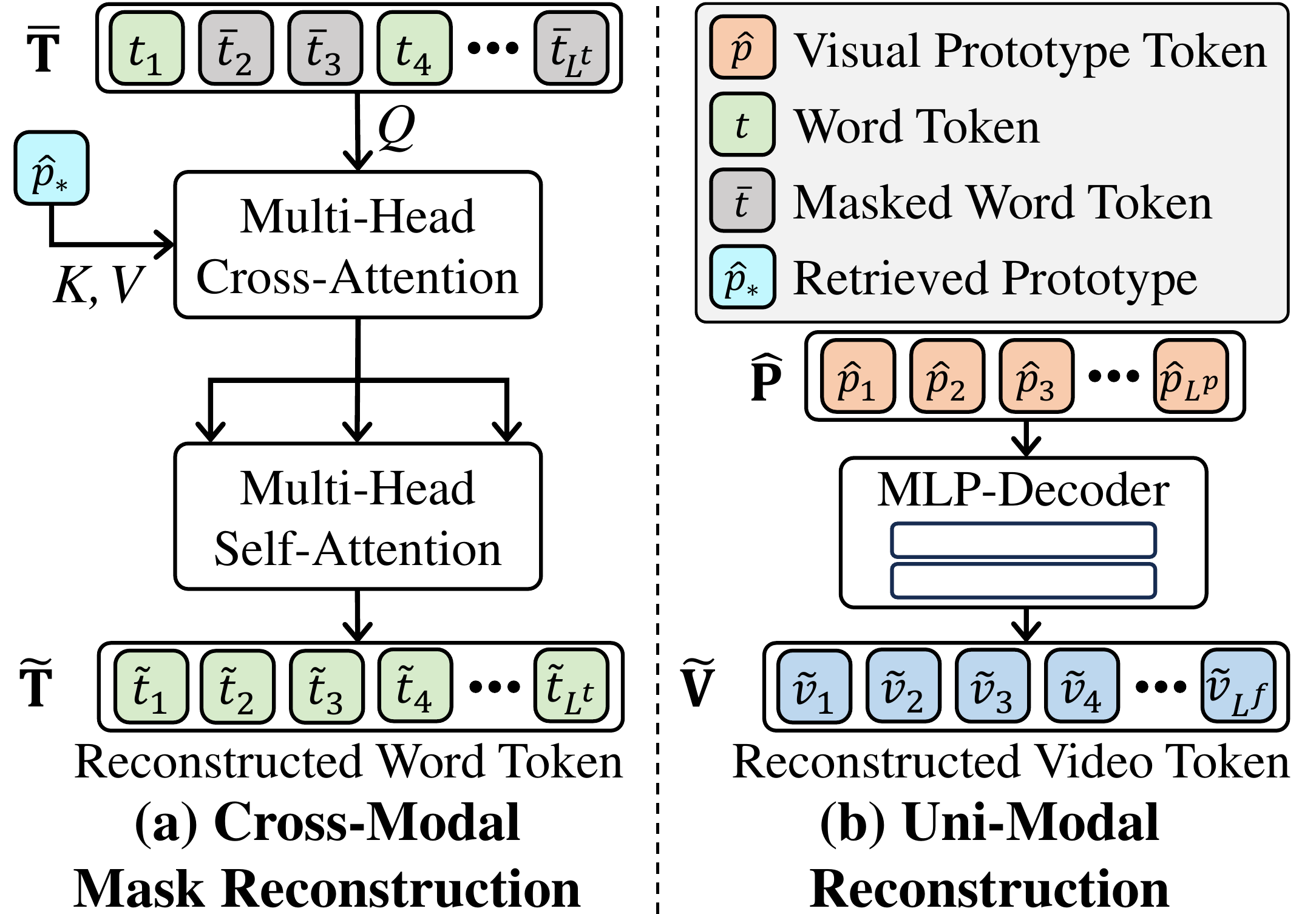}
    \vspace{-0.2cm}
    \caption{An architectural overview of reconstruction tasks.
    (a) For a cross-modal scenario, masked query features are reconstructed with the retrieved video prototype $\hat{p}_{*}$ via a transformer-based decoder.
    This aligns the retrieved prototype with the corresponding textual features.
    (b) For a uni-modal scenario, visual prototypes are processed via an MLP-based decoder to reconstruct the frame-wise features to mitigate the visual information loss during aggregation.
     }
    \label{fig:mask_recon}
    \vspace{-0.4cm}
\end{figure}

Our reconstruction tasks are two-fold: cross-modal and uni-modal.
While the cross-modal mask reconstruction promotes the alignment between the prototypes and the textual features, the uni-modal reconstruction ensures the preservation of visual contexts within the prototypes.
Reconstruction processes are depicted in Fig.~\ref{fig:mask_recon}.


\subsubsection{Cross-modal Mask Reconstruction: \\ Bridging the Modality Gaps}
The modality gap is one of the core challenges for effective PRVR. 
To enable the visual prototypes to better understand text queries, we propose predicting masked text tokens using the visual prototypes. 
However, using all visual prototypes to reconstruct specific queries might lead to representation collapse, as text queries contain only partial semantics of the video. 
To prevent this, we use only the retrieved visual prototype $\hat{p}_*$ to reconstruct the masked text features, as shown in Fig.~\ref{fig:mask_recon}~(left).
Masked text features are defined using a set $\mathcal{M}$, which specifies the indices of masked tokens, as follows:
\begin{gather}
    \bar{\mathbf{T}} = \left[ \acute{t}_i \mid i \in \{1, 2, ..., L^t\}, \, \acute{t}_i = \begin{cases} 
    \bar{t}_i, & \text{if} \, i \in \mathcal{M} \\
    t_i, & \text{if} \, i \notin \mathcal{M}
    \end{cases} \right],
\end{gather}
where $\bar{t}$ and $t$ each denote masked and unmasked text tokens.
Then, the cross-modal reconstruction is formulated as:
\vspace{-0.1cm}
\begin{gather}
\label{Eq.unimaskrecon}
    \tilde{\mathbf{T}} = \text{SA}\left(\text{CA}\left( \text{fc}_q( \bar{\mathbf{T}} ) \;; \text{fc}_k(\hat{p}_*)\;; \text{fc}_v(\hat{p}_*) \right)\right),
\end{gather}
where $\text{fc}_q$, $\text{fc}_k$, and $\text{fc}_v$ are to align the dimension between the masked input and the prototypes, and SA/CA denote the multi-head self-attention and cross-attention layers, respectively.
To clarify, the intuition behind using masked reconstruction instead of na\"ve reconstruction is that the same semantics can be expressed through diverse language formulations. 
Thus, we aim to minimize reconstruction errors caused by expression variations by providing structured references for reconstruction. 
Specifically, unmasked word tokens in each query serve as templates within the attention-based decoder, where attention mechanisms establish dependencies between them.
This ensures that the retrieved prototype accurately captures the semantics of the query, enabling the model to align prototypes with the given textual context.

The reconstruction objective for each text query is designed with infoNCE~\cite{simclr} against all other masked word tokens within the mini-batch.
Specifically, let $t^{b}_{j}$ represent the word tokens extracted from backbone, which corresponds to the $j$-th masked word in $b$-th text query in mini-batch.
Then, the loss for $i$-th reconstructed text word $\tilde{t}_{i}$ is expressed as:
\begin{equation}
    \!\mathcal{L}_{\text{crecon}}\! = \!-\frac{1}{\vert\mathcal{M}\vert}\!\!\sum\limits_{i\in \mathcal{M}}\!\text{log}\frac{\text{exp}(\tilde{t}_{i} \cdot t_{i}  )}
    {{\sum\limits_{b\in \mathcal{B}}}{\sum\limits_{j\in \mathcal{M}}} \! \text{exp}(\tilde{t}_{i}\! \cdot t^{b}_{j}  )},
\end{equation}
where $\mathcal{B}$ indicates the set of query indices within a batch.
We note that, for simplicity, we assume that the masked word indices are identical across all text queries within a batch, allowing $\mathcal{M}$ to be shared.
Cosine similarity is used as a metric since the goal of PRVR is to optimize the cosine distance for text-video retrieval.


\subsubsection{Uni-modal Reconstruction: Context Preservation}
Although there are benefits to text-video alignment from cross-modal task, cross-modal reconstruction may compromise visual context preservation by selectively focusing on the most text-associative visual details within the prototype.
To mitigate the information loss, we simultaneously perform the uni-modal reconstruction task, shown in Fig.~\ref{fig:mask_recon}~(right).

\begingroup
\setlength{\tabcolsep}{4.5pt} 
\renewcommand{\arraystretch}{0.9} 
\begin{table*}[t]
\vspace{-0.3cm}
\centering
\small
{ 
 \caption{Performance comparison on TVR and ActivityNet Captions dataset. Rows highlighted in gray represent T2VR and VCMR methods. Note that VCMR methods were trained without using moment annotations.}
\label{table_tvr}
\vspace{-0.2cm}
 \begin{tabular}{l|cccccc|cccccc}
 \hlineB{2.5}
 \multicolumn{1}{c|}{\multirow{2}{*}{Method}} & \multicolumn{6}{c|}{TVR} & \multicolumn{6}{c}{ActivityNet Captions} \\
\multicolumn{1}{l|}{}  & R@1 & R@5 & R@10 & R@100 & SumR & Memory~(MB) & R@1 & R@5 & R@10 & R@100 & SumR & Memory~(MB) \\ \hlineB{2.5}
\multicolumn{1}{c}{} & \multicolumn{12}{c}{ResNet152 + I3D + Roberta} \\ \hlineB{2.5}
\rowcolor{gray!40}
\rowcolor{gray!40}
CLIP4Clip & 9.9  & 24.3 & 34.3 & 72.5 & 141.0 & - & 5.9 & 19.3 & 30.4 & 71.6 & 127.3 & -\\
\rowcolor{gray!40}
Cap4Video & 10.3 & 26.4 & 36.8 & 74.0& 147.5& - & 6.3 & 20.4 & 30.9 & 72.6 & 130.2 & -\\
\rowcolor{gray!40}
XML  & 10.0& 26.5 & 37.3 & 81.3 & 155.1& - & 5.3 & 19.4& 30.6 & 73.1  & 128.4 & -\\
\rowcolor{gray!40}
ReLoCLNet & 10.7 & 28.1 & 38.1 & 80.3 & 157.1& - & 5.7 & 18.9& 30.0& 72.0 & 126.6 & - \\
\rowcolor{gray!40}
CONQUER& 11.0& 28.9 & 39.6 & 81.3 & 160.8& - & 6.5 & 20.4& 31.8 & 74.3  & 133.1 & - \\ \hline
MS-SL& 13.5 & 32.1 & 43.4 & 83.4 & 172.4 & 1.01 & 7.1 & 22.5& 34.7 & 75.8  & 140.1 & 0.98 \\
PEAN & 13.5 & 32.8 & 44.1 & 83.9 & 174.2 & - &  7.4 & 23.0 &35.5 &75.9 &141.8 & -\\
T-D3N & 13.8 & 33.8 &  45.0  & 83.9 &  176.5 & - &  7.3 &  23.8 &  36.0 &  76.6 &  143.6 & -\\
GMMFormer & 13.9 & 33.3 & 44.5 & 84.9 & 176.6&  0.05 & \textbf{8.3} & \textbf{24.9} & 36.7 & 76.1  & 146.0 & 0.05 \\ \hline
\rowcolor{gray!10}
\textbf{Ours} & \textbf{15.4} & \textbf{35.9} & \textbf{47.5} & \textbf{86.4} & \textbf{185.1} & 0.09 & 7.9 & \textbf{24.9} & \textbf{37.2} & \textbf{77.3} & \textbf{147.4} & 0.09 \\ \hlineB{2.5}
\multicolumn{1}{c}{} & \multicolumn{12}{c}{ResNet152 + I3D + Roberta + CLIP B/32} \\ \hlineB{2.5}
DL-DKD & 14.4 & 34.9 & 45.8 & 84.9 & 179.9 & 0.39 & 8.0& 25.0 & 37.4 &77.1  & 147.6 & 0.39 \\ \hlineB{2.5}
\multicolumn{1}{c}{} & \multicolumn{12}{c}{CLIP L/14} \\ \hlineB{2.5}
QASIR 2x2~ & 23.0 & 45.4 & 56.3 & 88.9 & 213.6 & 0.10 & - & - & - & -  & - & - \\ 
MS-SL & 30.8 & 57.1 & 67.1 & 93.3 & 248.2 & 1.01 & 14.6 & 37.1 & 50.2 & 84.4 & 186.3 & 0.99  \\ 
GMMFormer & 29.4 & 56.8 & 65.5 & 93.3 & 243.2 & 0.05 & 15.0 & 37.6 & 50.3 & 83.9 & 186.9 & 0.05 \\ \hline
\rowcolor{gray!10}
\textbf{Ours} & \textbf{34.7} & \textbf{60.0} & \textbf{70.1} & \textbf{94.4} & \textbf{259.2} & 0.09 & \textbf{16.0} & \textbf{38.8} & \textbf{52.4} & \textbf{85.1} & \textbf{192.3} & 0.09 \\ \hlineB{2.5}
 \end{tabular}
 \vspace{-0.25cm}
}
\end{table*}
\endgroup

Uni-modal encoding ensures that the contexts of the whole video are preserved within the visual prototypes~\cite{slot}.
Particularly, the encoded visual prototypes $\hat{\mathbf{P}}\in \mathbb{R}^{L^P \times D}$ are decoded through an MLP-based decoder and broadcasted to the $L^f$ dimension to reconstruct the semantics of the input visual features $\mathbf{V}$.
Given that the reconstructed outputs are $\tilde{\mathbf{V}}=\!\left[\tilde{v}_{i} | i\in\{1, 2, ..., L^f\}\right]$,
L2 distance is employed to calculate the error of the predicted mask tokens as follows:
\vspace{-0.2cm}
\begin{equation}
\label{Eq.unimaskreconloss}
\mathcal{L}_{\text{urecon}} = \frac{1}{L^{f} } \sum\nolimits_{i=1}^{L^f}
    \left( v_{i} - \tilde{v}_{i} \right)^2.
\end{equation}

\noindent Finally, the loss for dual reconstruction is formed as the summation of $\mathcal{L}_{\text{crecon}}$ and $\mathcal{L}_{\text{urecon}}$:
\begin{equation}
    \mathcal{L}_{\text{recon}} = \mathcal{L}_{\text{crecon}} + \mathcal{L}_{\text{urecon}}.
\end{equation}

\subsection{Weak Guidance for Where to Focus}
\label{sec.guidance}
In prototypical retrieval, it is crucial for the retrieved prototypes to capture the visual contexts relevant to the given text query.
However, in PRVR, constraining prototypes to encode only the contextually relevant segments is challenging due to the absence of dense temporal span annotations.

To circumvent this limitation, we provide weak supervision for where to focus by concatenating two different videos. 
To be specific, we guide the model's attention weight with the concatenated examples by training the model to focus on the video that corresponds to each query.
To determine which video the retrieved prototype is attending to, we compute the total attention weights of the retrieved prototype for each video.
Let us denote the sum of attention weights for each video as the scalars $\dot{\alpha}_1$ and $\dot{\alpha}_2$.
Given that the sum of $\dot{\alpha}_1$ and $\dot{\alpha}_2$ equates to 1, we apply a binary cross-entropy loss on only $\dot{\alpha}_1$ with the label $L$ which is assigned 1 if the first video of concatenated instance is relevant to the query and otherwise 0.
The objective is formulated as follows:
\begin{equation}
\!\!\!\!\mathcal{L}_{\text{attn}}\! =\! -\left[L \log(\dot{\alpha}_1 + \beta)\! +\! (1 - L) \log(1 - (\dot{\alpha}_1 \!- \beta))\right]\!,
\end{equation}
where $\beta$ is margin to mitigate the overfitting in case contextually similar videos are mixed.
Although this method does not provide explicit signals for the model to learn the exact video frames to attend to, it indirectly discourages the model from considering completely irrelevant video segments.



\subsection{Prototype Diversification}
\label{sec.diversification}
Despite the benefits of prototypical framework, the capability to generalize across diverse text queries may be limited if semantic overlap exists between prototypes within the same video.
To address this, we incorporate an orthogonal objective into prototypical learning to promote prototype diversification.
Formally, the orthogonal loss within the prototypes is expressed as:
\begin{equation}
\label{Eq.ortho}
    \mathcal{L}_{\text{ortho}}=\frac{1}{L^p\left(L^p-1\right)} \vert \vert \mathbf{O} - \mathbf{I}_{L^p} \vert\vert ; \mathbf{O} = \text{max}(\mathbf{\hat{P}} \mathbf{\hat{P}}^T, 0),
\end{equation}
where $\textbf{I}_{L^p}$ is $L^p \times L^p$ identity matrix.
Note that prototype separation is applied only when their cosine similarity is positive. 
This prevents overly strict regularization and allows for partial semantic overlap, acknowledging that some videos may contain only a limited range of events.

\noindent\textbf{Overall Objective.} \;Finally, our objective is:
\begin{equation}
\mathcal{L} = \sum\nolimits_{x \in \{\text{ret}, \text{recon}, \text{attn}, \text{ortho}\}} \lambda_x \mathcal{L}_{x},
\vspace{-0.05cm}
\end{equation}
where $\lambda_x$ represents the loss coefficient specific to each $\mathcal{L}_{x}$ and $\mathcal{L}_{\text{ret}}$ denotes the standard retrieval loss.  
Note that these coefficients are mostly set consistent across datasets except for $\mathcal{L}_\text{ret}$ following previous work~\cite{gmmformer} and orthogonal loss to reflect the characteristics of datasets~(details in Appendix).



\section{Experiments}
\label{sec:experiment}


\noindent\textbf{Datasets \& Backbones.}
We use two widely used large-scale datasets, TVR~\cite{tvr} and ActivityNet Captions~\cite{anetcaptions}. 
Also, we re-organize QVHighlights~\cite{qvhighlights}, a moment retrieval dataset, for PRVR evaluation.
Details are in the Appendix.
For experiments on TVR and AcitivityNet Captions, we utilize two different backbones.
For the settings with ResNet152, I3D, and Roberta, we adopt the settings from MS-SL~\cite{prvr}.
Also, beyond conventional benchmarks, we compare methods with a stronger backbone, CLIP-L/14~\cite{CLIP} following QASIR~\cite{qasir}. 
For a comparison of backbone adequacy for PRVR, we refer to the Appendix, where we highlight that CLIP extracts more distinguishable representations for retrieval.
Note that CLIP features are extracted at 3 fps for all datasets, and official implementations of recent works are used to reproduce performances with CLIP-L/14.
Slowfast~\cite{feichtenhofer2019slowfast} and CLIP-B/16 are used for QVHighlights following its original work~\cite{momentdetr}.

\noindent\textbf{Evaluation.} 
Following \cite{prvr}, the rank-based metrics \textit{R@M} ($M=1,5,10,100$) are reported to measure the retrieval accuracy.
\textit{R@M} is defined as the ratio of accurate retrievals given queries, where the retrieval is considered accurate if the correctly corresponding video is included within the top M of the retrieved list.
For overall performance comparison, we use the Sum of all Recalls (SumR).



\begin{figure}[t]
\vspace{-0.3cm}
    \centering
    \includegraphics[width=0.97\columnwidth]{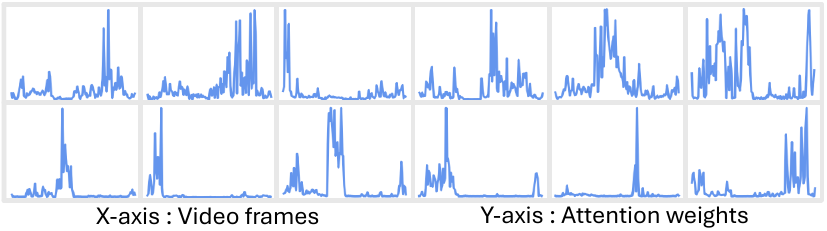}
    \vspace{-0.25cm}
    \caption{
        An example of prototypes' attention weights on each frame.
        Each box represents each prototype.
    }
    \label{fig:instance_attn}
    \vspace{-0.35cm}
\end{figure}

\subsection{Comparison with the State-of-the-arts}
Results are in Tab.~\ref{table_tvr} against state-of-the-art~(SOTA) PRVR methods~\cite{prvr, gmmformer, cheng2024transferable, pean, dldkd, qasir} along with T2VR~\cite{clip4clip, cap4video}, and VCMR methods without moment annotations~\cite{reloclnet, tvr, conquer}. 
On TVR, our proposed method outperforms the SOTA techniques by a large margin up to 8.5 in SumR using a combination of ResNet152, I3D, and Roberta, and 11 in SumR with CLIP-L/14.
Consistent results are also observed with the ActivityNet Captions dataset, where our method continues to outperform existing SOTA approaches.
We attribute this enhanced performance to the prototypes' effectiveness in preserving and managing diverse video contexts.
In particular, previous methods typically select only a single segment per video~(the one with the highest similarity to the text query) for training text-video retrieval. 
This approach limits the model to learning from a single representative element per video, potentially excluding other video segments from model learning.
In contrast, our prototypes are constructed as mixtures of multiple video segments, as shown in Fig.~\ref{fig:instance_attn}. 
This allows the model to incorporate a broader range of visual representations, mitigating the issue of learning from only a single video segment.
Also, we identify specific challenges associated with GMMFormer~\cite{gmmformer}, particularly when utilizing a context-rich backbone such as CLIP.
We attribute this issue to the local-biased encoding strategy of GMMFormer, which derives limited benefit from a powerful CLIP backbone.
Our findings indicate that encoding all possible cases without constraints better exploits the powerful representation extracted from foundation backbones~(CLIP).

\begingroup
\setlength{\tabcolsep}{4.8pt} 
\renewcommand{\arraystretch}{0.9} 
\begin{table}[t]
{
 \centering
 \small
 \vspace{-0.2cm}
 \caption{Results on QVHighlights $val$ split. 
 }
 \vspace{-0.2cm}
\label{table_qv}
 \begin{tabular}{l|ccccc}
 \hlineB{2.5}
Model  & R@1 & R@5 & R@10 & R@100 & SumR  \\ \hline
MS-SL~\cite{prvr}& 20.4 & 46.7 & 60.7 & \textbf{94.6} & 222.5 \\
GMMFormer~\cite{gmmformer} & 18.2 & 43.7 & 56.7 & 92.5 & 211.1 \\ 
\rowcolor{gray!10}
\textbf{Ours} & \textbf{22.6} & \textbf{48.8} & \textbf{61.3} & 93.9 & \textbf{226.6} \\ \hline
 \end{tabular}
 \vspace{-0.25cm}
}
\end{table}
\endgroup

Also, our method demonstrates notable efficiency in storage requirements by reducing memory consumption by 90\% compared to MS-SL, which similarly leverages diverse contextual information of varying lengths. 
Note that GMMFormer's high efficiency in storage is because they only store a single video-level clip along with local clips, at the expense of encoding diverse contexts. 
In contrast, our approach prioritizes maintaining rich contextual information while remaining efficient.
Indeed, our study for the trade-off between the performance and memory consumption in Tab.~\ref{table_num_proto} validates the benefits of our prototypical framework; SOTA performance is achieved even when less memory is consumed by reducing the number of prototypes.
\begingroup
\setlength{\tabcolsep}{5.1pt} 
\renewcommand{\arraystretch}{0.9} 
\begin{table}
\footnotesize
\vspace{-0.05cm}
\caption{Inference time~(ms) and matching FLOPs~(G) comparison for each text query on TVR. Matching FLOPs in the parentheses refer to FLOPs for video-text matching process.}
\label{table_time}
\vspace{-0.2cm}
\centering
{
\begin{tabular}{l|cccc}
\hlineB{2.5}
Video Size & 1000 & 2000 & 3000 & 4000 \\ \hline
MS-SL & 0.53~(0.50) & 0.91~(1.01) & 1.25~(1.51) & 1.66~(2.02) \\
GMMFormer & 0.29~(0.03) & 0.31~(0.05) & 0.32~(0.08) & 0.33~(0.10) \\ 
\rowcolor{gray!10}
Ours & 0.39~(0.05) & 0.39~(0.09) & 0.43~(0.14) & 0.47~(0.18) \\ \hlineB{2.5}
\end{tabular}
}
\vspace{-0.4cm}
\end{table}
\endgroup











For further validation, we re-organize QVHighlights as a new PRVR dataset.
In Tab.~\ref{table_qv}, we present a comparison against recent methods.
As shown, our method outperforms the baselines, achieving an improvement of over 2\%p in the R@1 metric. 
In contrast, we find a similar trend in GMMFormer that it yields limited performances when attached to a powerful vision-language model, CLIP.


Lastly, we report computation efficiency in Tab.~\ref{table_time}.
Particularly, inference time and FLOPs during the text-video retrieval process are measured across various video database sizes using an NVIDIA RTX 3090 GPU and an Intel Xeon Gold 5220R CPU. 
Results indicate that our proposed method's efficiency is comparable to that of GMMFormer, while significantly outperforming MS-SL in terms of speed as the video storage size increases.

\subsection{Analysis}
\label{sec.ablation}
We note all studies were conducted on TVR using CLIP.

\begingroup
\setlength{\tabcolsep}{2.3pt} 
\renewcommand{\arraystretch}{0.9} 
\begin{table}[t]
\vspace{-0.3cm}
\centering
{
 \footnotesize
 \caption{
     Ablation study on model components.
     From left to right, we sequentially add cross-modal mask reconstruction, uni-modal reconstruction, weak guidance for prototype, and prototype diversification objective. 
 }
 \vspace{-0.3cm}
 \label{table_ablation}
 \begin{tabular}{cccc|ccccc}
 \hlineB{2.5}
 C.Recon & U.Recon & Guide & P.Div. & R@1 & R@5 & R@10 & R@100 & SumR \\ \hline
 - & - & - & - & 32.8 & 58.2 & 68.6 & 93.4 & 253.0  \\ 
 \checkmark & - & - & - & 32.9 & 58.7 & 69.0 & 93.5 & 254.2  \\ 
 \checkmark & \checkmark & - & - & 33.8 & 58.7 & 68.8 & 93.9 & 255.1 \\
 \checkmark & \checkmark & \checkmark & - & 34.3 & 60.0 & 69.9 & 94.1 & 258.2  \\ 
 \checkmark & \checkmark & \checkmark & \checkmark & 34.7 & 60.0 & 70.1 & 94.4 & 259.2 \\ 
 \hlineB{2.5}
 \end{tabular}
 \vspace{-0.05cm}
}
\end{table}
\endgroup



\noindent\textbf{Component Ablation.} In Tab.~\ref{table_ablation}, we report an component study.
On top of our baseline only with the prototype generator, we progressively introduce and evaluate the effects of (1) cross-modal mask reconstruction, (2) uni-modal reconstruction, (3) weak attention guidance, and (4) prototype orthogonal objective.
Our finding indicates that the prototypical network for PRVR establishes a solid baseline, as demonstrated in the first row.
This implies that the implicit prototype aggregation with exhaustive clip modeling is a powerful approach for encoding semantically meaningful contexts, especially with a strong CLIP-L/14 backbone.
Furthermore, the results from the second row to the last row validate that each proposed component effectively complements the implicit nature of prototypical learning, enhancing the overall retrieval performance.

\begingroup
\setlength{\tabcolsep}{11.5pt} 
\renewcommand{\arraystretch}{0.92} 
\begin{table} 
\vspace{-0.2cm}
\caption{Performances w.r.t. the number of prototypes.}
\label{table_num_proto}
\vspace{-0.2cm}
\centering
\footnotesize
{
\begin{tabular}{l|cccc}
\hlineB{2.5}
Num of Proto. & 10 & 20 & 30 & 60 \\ \hline
SumR & 252.9 & 257.8 & 259.2 & 260.2  \\
Memory~(MB) & 0.03 & 0.06 & 0.09 & 0.18   \\ 
\hlineB{2.5}
\end{tabular}
\vspace{-0.05cm}
}
\end{table}
\endgroup

\begingroup
\setlength{\tabcolsep}{4.4pt} 
\renewcommand{\arraystretch}{0.9} 
\begin{table}[t]
\centering
{
 \small
 \caption{
    Performance comparison with different architectural designs for prototype construction.
 }
 \label{table_perceiver_slot}
 \vspace{-0.25cm}
 \centering
 \footnotesize
 \begin{tabular}{c|c|ccccc}
 \hlineB{2.5}
 & Model & R@1 & R@5 & R@10 & R@100 & SumR \\ \hline
 (a) & Perceiver & 31.8 & 57.1 & 67.0 & 93.5 & 249.5~(-9.7) \\ 
 (b) & Slot-attention & 32.3 & 58.4 & 68.9 & 94.1 & 253.8~(-5.4) \\ 
 \rowcolor{gray!10}
 (c) & Ours & 34.7 & 60.0 & 70.1 & 94.4 & 259.2~($\pm$0.0) \\ 
 \hlineB{2.5}
 \end{tabular}
 \vspace{-0.3cm}
}
\end{table}
\endgroup

\noindent\textbf{Varying Number of Prototypes.}
Number of the prototypes is a critical parameter that balances the trade-off between performance and memory consumption~(which also directly affects time efficiency).
In Tab.~\ref{table_num_proto}, our analysis explores the impact of the number of prototypes.
Our observation reveals that a small number of prototypes yields the highest efficiency albeit at the cost of reduced performance.
Nevertheless, we claim that using only 10 prototypes already achieves strong performance, offering a more efficient and effective approach compared to our baseline, GMMFormer~(which achieves a SumR of 243.2 with a memory requirement of 0.05MB).
On the other hand, an increased number of prototypes to 60 improves the performance but incurs a linear increase in memory consumption. 
Considering the trade-off between performance and efficiency, we propose that using 30 prototypes strikes an optimal balance between the two.

\noindent\textbf{Architecture for Prototype Construction.}
As discussed in Sec.~\ref{Sec.prototypeconstruction}, our prototype generator is deliberately designed using only cross-attention. 
In this study, we compare our architecture with more complex designs that are widely adopted in the literature~\cite{jaegle2021perceiver, slot}. 
Notably, the Perceiver and Slot-attention architectures incorporate additional components such as self-attention and Gated Recurrent Unit~(GRU), which can introduce unnecessary interactions between prototypes and increase computational overhead.
Results in Tab.~\ref{table_perceiver_slot} underscore the effectiveness of our streamlined design for PRVR, demonstrating its superiority in performance.

\begin{figure}[t!]
    \vspace{-0.3cm}
    \centering
    \includegraphics[width=0.48\textwidth]{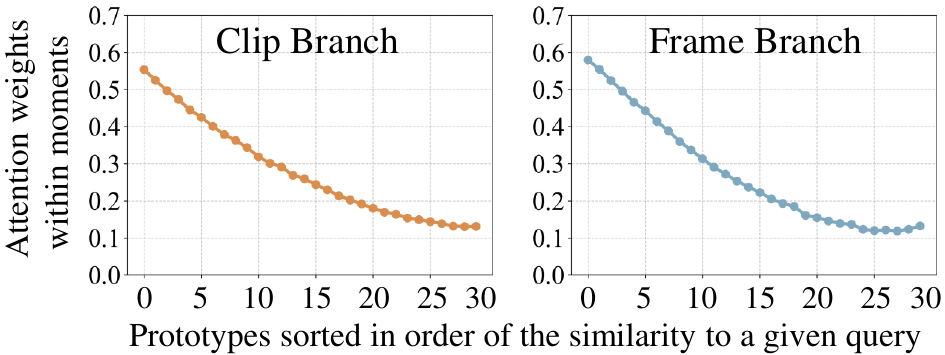}
    \vspace{-0.37cm}
    \caption{Attendance on frames within moments for visual prototypes in the order of the similarity to a given query.
    The higher the similarity between the prototype and the text query, we observe the higher the attendance on moment frames.
    }
    \label{fig:attention_qualitative_rank}
    \vspace{-0.1cm}
\end{figure}

\begin{figure}[t!]
\vspace{-0.2cm}
    \centering
    \includegraphics[width=0.48\textwidth]{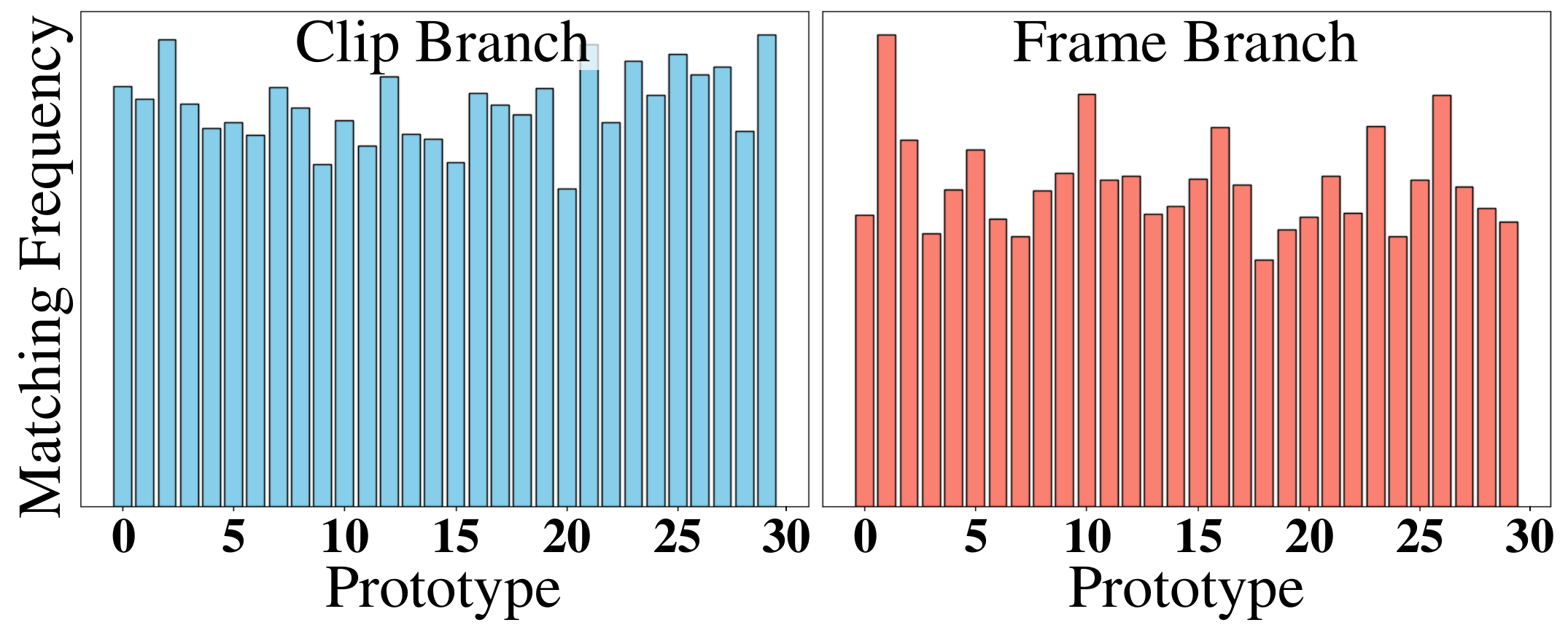}
    \vspace{-0.6cm}
    \caption{
    Matching frequency of prototypes on TVR dataset.
    }
    \label{fig:frequency}
    \vspace{-0.1cm}
\end{figure}
\begin{figure}[t!]
\vspace{-0.1cm}
    \centering
    \includegraphics[width=0.8\columnwidth]{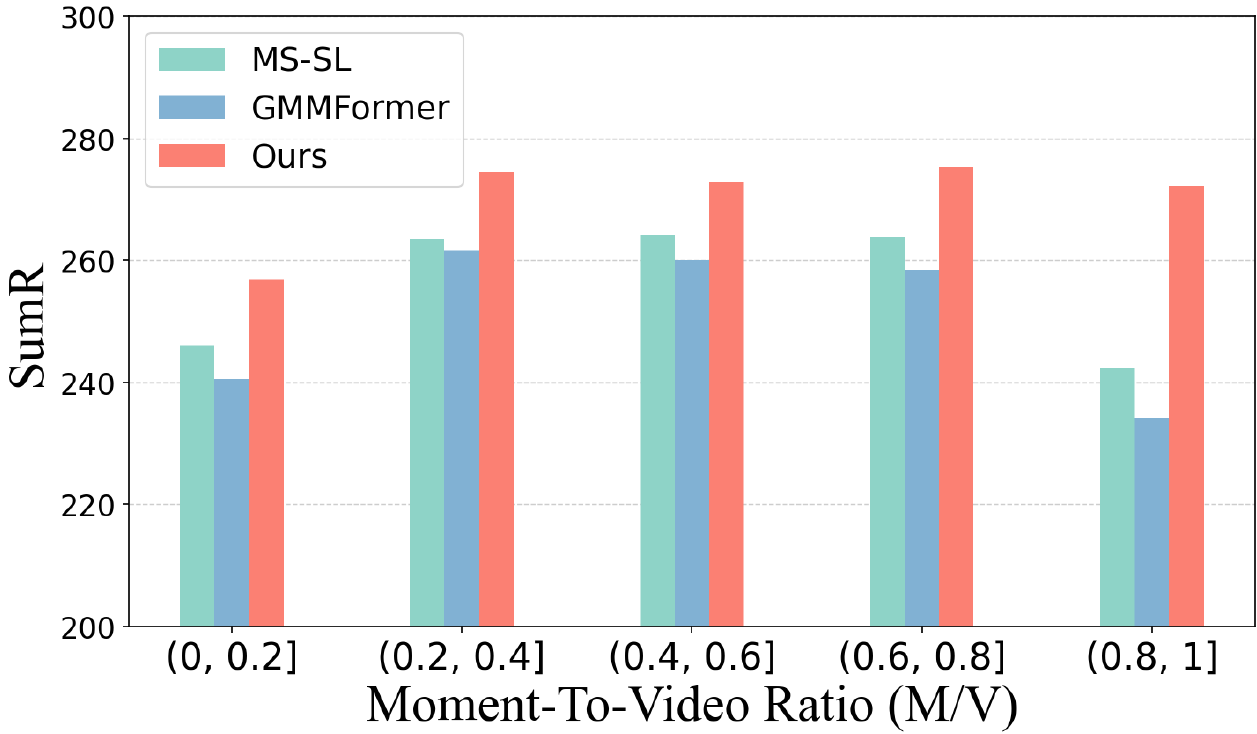}
    \vspace{-0.24cm}
    \caption{
    Performances \textit{w.r.t.} the ratio of moments in videos.
    }
    \label{fig:mtvratio}
    \vspace{-0.3cm}
\end{figure}

\begin{figure*}[t!]
    \vspace{-0.3cm}
    \centering
    \includegraphics[width=0.93\textwidth]{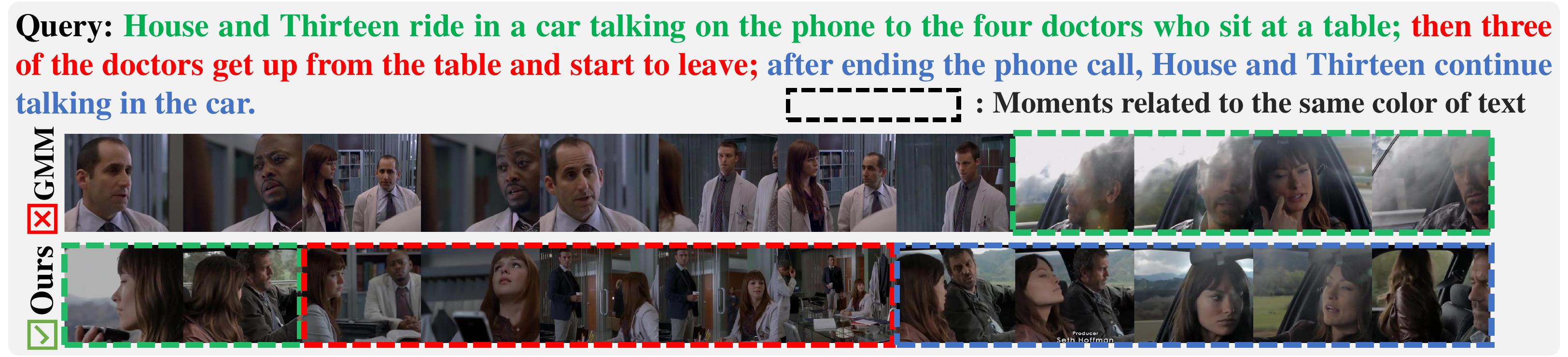}    
    \vspace{-0.15cm}
    \caption{Qualitative results on TVR dataset. 
    Each context in the query is colored in different colors while the corresponding moments are marked with the same color if the context exists in the video.
    }
    \label{fig:long_moment_qualitative}
    \vspace{-0.1cm}
\end{figure*}
\begin{figure*}[t!]
\vspace{-0.2cm}
    \centering
    \includegraphics[width=0.93\textwidth]{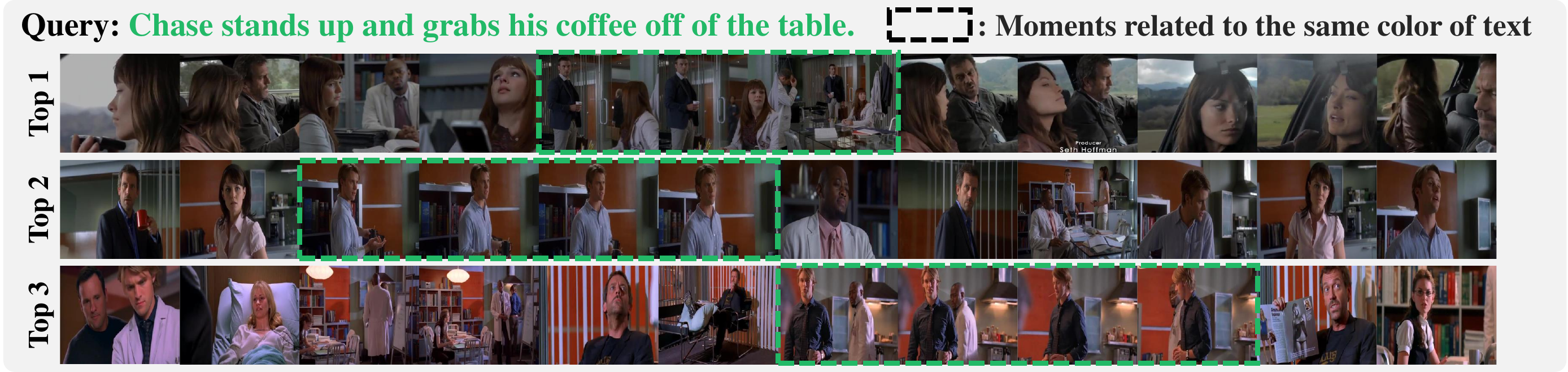}    
    \vspace{-0.15cm}
    \caption{Examples of top-ranked retrieved videos on TVR. Moments where the query context is present are highlighted with a dotted line.
    }
    \label{fig:top3_qualitative}
    \vspace{-0.35cm}
\end{figure*}

\noindent\textbf{Attention Analysis of Prototypes.}
We analyze whether prototypes encode text-relevant semantics.
Particularly, we visualize the magnitude of prototypes' attention weights on the in-moment frames that include the text-relevant contexts.
Note that the temporal margin is added to moment boundaries~(i.e., start and end indices of moments are formed as [st-tm, ed+tm] following \cite{zhao2017temporal, lin2018bsn}) since understanding nearby context is crucial to comprehend the moments~(tm is set to 1/20th of moment's duration).
In Fig.~\ref{fig:attention_qualitative_rank}, we plot the attention magnitude within the moment frames of each prototype in the order of the similarity between each prototype and the text description.
Specifically, we observe that retrieval score~(similarity) is highly correlated with the degree of attendance on in-moment frames.
This validates the contextual understanding capability of our prototypical framework.

\noindent\textbf{Matching Frequency of Prototypes.}
To validate the utility and effectiveness of each prototype in the retrieval process, we present the matching frequency of prototypes in Fig.~\ref{fig:frequency}. 
This analysis shows the number of times each prototype is retrieved. 
A uniformly distributed frequency across prototypes indicates that all prototypes are actively contributing to the retrieval process, thereby confirming their effectiveness in representing the diverse content of the dataset.

\noindent\textbf{Robustness Across Moment Lengths.}
Our performance across varying moment lengths is also examined in Fig.~\ref{fig:mtvratio}.  
The results demonstrate the robustness of our prototypical framework in handling moments of different durations. 
We attribute this to the nature of visual prototypes, which aggregate visual context based on similarity, allowing both long and short events to be considered evenly. 
In contrast, our observations indicate that the baseline models struggle with longer moments. 
Specifically, for MS-SL, this challenge arises because shorter clips dominate the modeled segments. 
Similarly, GMMFormer has difficulty with longer moments due to its constraints to focus more on shorter moments.

\subsection{Qualitative Results}
Fig.~\ref{fig:long_moment_qualitative} shows retrieval results of GMMFormer and ours.
In this example, the query describes multiple sequential events, necessitating the model's ability to comprehend long, complex scenarios to retrieve the correct video.
Our method successfully addresses this challenge by effectively handling the complex query.
In contrast, GMMFormer tends to retrieve incorrect videos, focusing on only partial context as the most similar match.
This limitation arises because they primarily construct clip contexts only within the adjacent frames.
This comparison highlights the advantages of our prototypical framework; it is effective in recognizing long sequences of events and adapts to varying context lengths, leveraging the benefits of exhaustive clip modeling.

In Fig.~\ref{fig:top3_qualitative}, we illustrate how our prototypical framework interprets given context by examining top-3 retrieved videos.
Notably, we observe that these top-ranked videos exhibit a highly similar contextual alignment; the top-1 video thoroughly captures the intended context while top-2 and top-3 videos depict scenes where ``Chase is standing and holding his coffee".
This example highlights the capability of prototypes to learn and represent consistent contextual features.

\section{Conclusion}
This work addresses a key challenge often faced by retrieval systems: balancing retrieval accuracy and inference speed. 
To overcome this tradeoff, we introduce a novel prototypical learning framework that leverages rich contextual information of an exhaustive clip modeling strategy without compromising efficiency.
In particular, our prototypes efficiently aggregate features across diverse temporal scales into a compact representation.
Then, cross- and uni-modal reconstruction tasks are proposed to enhance the capability of text comprehension within the prototypes while ensuring that the prototypes effectively encode video contexts.
Also, we provide weak guidance and orthogonal loss to guide the retrieved prototypes in attending to the appropriate content and encourage the diversity between prototypes.
Extensive experiments demonstrate that our approach effectively balances the tradeoff between retrieval accuracy and efficiency.

\maketitlesupplementary

\section{Datasets}
To benchmark PRVR methods, we employ two popularly used large-scale datasets~(i.e., TVR and AcitivtyNet Captions), which contain natural language descriptions for multiple subsets of each video.
\textbf{TVR}~\cite{tvr} comprises 19,524 videos each with 5 text descriptions collected from 5 different TV shows.
For the data split, we follow previous works to allocate 87,175 and 17,435 moment pairs for training and testing, respectively.
\textbf{ActivityNet Captions}~\cite{anetcaptions} has 9,043 and 4,430 YouTube videos, with 33,721 and 15,753 text sentences for training and testing, respectively.
On average, 3.7 descriptions are provided per video.
\textbf{QVHighlights} is initially introduced as a moment retrieval dataset~\cite{momentdetr}.
It generally contains user-created lifestyle vlog videos on YouTube, captured via different devices, \textit{e.g.}, smartphones or GoPro, with different angles, \textit{e.g.}, first-person or third-person.
Particularly, we use the training and validation sets of QVHighlights for evaluating PRVR since dense annotations~(possibly needed for future analysis) are not available for the test set.
Therefore, QVHighlights for PRVR consists of 7,218 moment-text pairs in the training set and 1,550 pairs in the test set. 
To align with the partially relevant video retrieval~(PRVR) task, we reorganized the dataset by merging videos originating from the same source, resulting in an average of 3.3 text queries per video. 
Consequently, QVHighlights for the PRVR contains 2,214 videos in the training set and 474 videos in the test set while preserving the original 7,218 and 1,550 text queries, respectively.
The dataset used in this study can be accessed at \footnote{https://github.com/wjun0830/ProtoPRVR}.
Finally, \textbf{Charades} is not used in this work since it is less suitable for evaluating PRVR.
To illustrate, Charades primarily consists of broad text queries that often share similar contextual meanings with queries from other videos. 
For example, the query `\textit{the person opens the door}'~(video index: HQ8BB) conveys the same semantics as `\textit{a person opens the door}'~(video index: 3W1GP), `\textit{person opens the door}'~(video index: 8O07M), `\textit{a person opens a door}'~(video index: LLTBQ), `\textit{person opens the door}'~(video index: IUETR), and many others.
From a qualitative perspective, the average length of text queries is 6.2 words, which is significantly shorter than TVR~(12.2 words), ActivityNet Captions~(13.6 words), and QVHighlights~(10.5 words).
Additionally, the overall cosine similarity between text query representations extracted with CLIP~\cite{CLIP} is 0.84, much higher than in TVR~(0.56), ActivityNet~(0.68), and QVHighlights~(0.55), indicating considerable redundancy in text queries.
Given these limitations, we newly introduced QVHighlights to use for evaluating PRVR instead of Charades.

\section{Backbone Comparison}
In this subsection, we compare the adequacy of different backbones used for benchmarking methods on TVR and ActivityNet Captions datasets, with a particular focus on the text modality. 
Tab.~\ref{supp_table_textsimilarity} presents the statistics on the pairwise cosine similarity between text representations across different datasets, using Roberta and CLIP models.
Our findings reveal that the text-vision aligned model, CLIP, effectively captures the distinct characteristics of text annotations in large-scale datasets. 
In contrast, text features derived from the Roberta model exhibit a high degree of similarity~(an average cosine similarity among all text queries is 0.961), which may limit the evaluation protocol in fairly assessing the generalization capability of different models.
We hope that this insight will encourage future research to carefully consider the backbone for model evaluation for PRVR.
\begingroup
\setlength{\tabcolsep}{13pt} 
\renewcommand{\arraystretch}{1} 
\begin{table}[h]
\centering
{
 \small
 \caption{Similarity statistics of text query representations across datasets with different backbones. 
 We report the maximum, mean, median, and minimum values among all pairwise similarities.
 ANet indicates ActivityNet Captions dataset.}
 \label{supp_table_textsimilarity}
        \begin{tabular}{c|c|c|c|c}
        \hlineB{2.5}
        & \multicolumn{2}{c|}{CLIP} & \multicolumn{2}{c}{I3D+Roberta} \\ \hline
        & TVR & ANet & TVR & ANet \\ \hline
        max & 1.000 & 1.000 & 1.000 & 1.000 \\
        mean & 0.558 & 0.678 & 0.961 & 0.994 \\
        median & 0.553 & 0.683 & 0.961 & 0.994 \\
        min & 0.218 & 0.245 & 0.877 & 0.974 \\
        \hlineB{2.5}
        \end{tabular}
}
\end{table}
\endgroup

\begin{figure*}[t]
    \centering
    \includegraphics[width=0.8\textwidth]{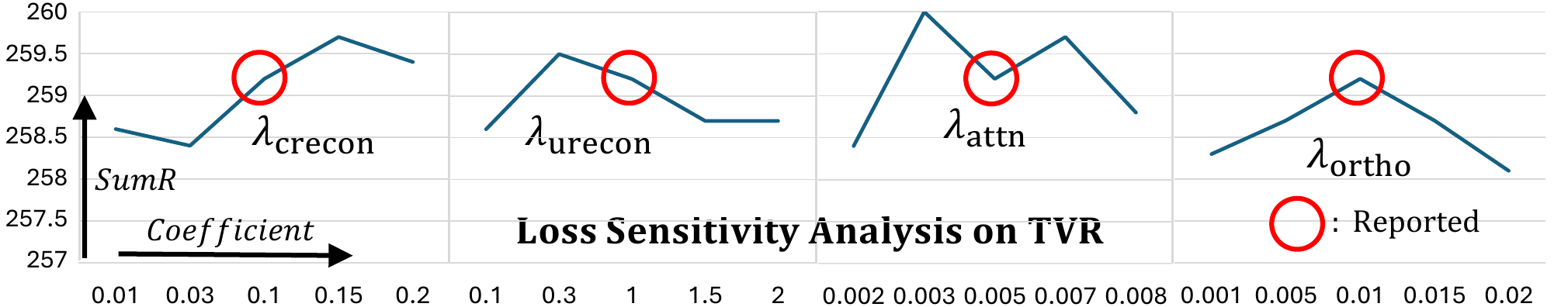}
    \caption{
        Studies on loss coefficients. 
        From left to right, each plot examines the sensitivity of $\lambda_\text{attn}$, $\lambda_\text{ortho}$, $\lambda_\text{crecon}$, and $\lambda_\text{urecon}$.
        In each plot, the X-axis represents the coefficient values, while the Y-axis indicates performance. 
        The red circles denote the default value.
     }
    \label{fig:loss_param}
\end{figure*}

\section{Implementation Details}
\label{sec:sup_implementation}

Hyperparameters used for the TVR, ActivityNet Captions, and QVHighlights datasets are listed in Tab.~\ref{supp_table_hyperparameter}. 
While most hyperparameters were kept consistent across the datasets, we adjusted the coefficients for contrastive loss~($\lambda_\text{nce}$) following \cite{gmmformer} to reflect the different characteristics of TVR dataset which contains proper nouns and similar contexts across text-video pairs since the dataset is collected from TV show.
\begingroup
\setlength{\tabcolsep}{13pt} 
\renewcommand{\arraystretch}{1} 
\begin{table}[h]
\centering
{
 \small
 \caption{Implementation details. From top to bottom, '$\lambda_\text{*}$' denotes the coefficient for corresponding losses~(i.e., contrastive~(InfoNCE) loss, cross-modal reconstruction loss, uni-modal reconstruction loss, attention guidance loss, and orthogonal loss within the prototypes); $\beta$ indicates the margin in attention guidance loss; $K$ stands for number of iterations for constructing prototypes.
 ANet and QVH denote ActivityNet Captions and QVHighlights, respectively.
 }
 \label{supp_table_hyperparameter}
        \begin{tabular}{c|c|c|c}
        \hlineB{2.5}
        & TVR & ANet & QVH \\ \hline
        $\lambda_\text{nce}$ & 0.03 & 0.06 & 0.06\\
        $\lambda_\text{crecon}$ & 0.1 & 0.1 & 0.1 \\
        $\lambda_\text{urecon}$ & 1.0 & 1.0 & 1.0\\
        $\lambda_\text{attn}$ & 0.005 & 0.005 & 0.005 \\
        $\lambda_\text{ortho}$ & 0.01 & 0.01 & 0.01 \\
        $\beta$ & 0.2 & 0.2 & 0.2 \\
        $K$ & 1 & 1 & 1\\
        \hlineB{2.5}
        \end{tabular}
}
\end{table}
\endgroup 

Furthermore, there are minor differences in implementation when employing the Roberta text encoder alongside CNN-based visual encoders. 
Specifically, we set $K$, the number of iterations for prototype construction, to 6 when using Roberta text encoder, and also assign higher value~(0.3) to $\lambda_\text{ortho}$ for experiments on ActivityNet Captions with Roberta.
This is due to the exceptionally high average similarity between text queries, particularly AcitivityNet Captions~(TVR: 0.961, ANet: 0.994), as shown in Tab.~\ref{supp_table_textsimilarity}.
Therefore, we increased $K$ to encourage the formation of more context-specific prototypes~(promoting greater distinction between video prototypes) and set a higher $\lambda_\text{ortho}$ to better separate the text queries.
Other than those, we kept the parameters the same.


For general training, we used the PyTorch framework for implementation and trained the models on NVIDIA RTX A6000 GPU with a batch size of 128 following previous works~\cite{prvr, gmmformer}.

\section{Loss Sensitivity} 
Fig.~\ref{fig:loss_param} presents our study on loss sensitivity, demonstrating the robustness of our prototypical framework as long as the loss coefficients remain within a reasonable range.
Furthermore, we remark that we maintain uniform loss coefficients across all datasets to validate the generalizability of our prototypical framework, regardless of domain-specific variations.
As a result, our default setting may not always achieve optimal performance across all datasets.
However, we argue that adjusting the loss weights for each dataset can further enhance performance, indicating that our approach can adaptively bring more gains from dataset-specific optimizations.

\vspace{5pt}
\noindent\textbf{Acknowledgements.} 
This project was partly supported by the NAVER Cloud Corporation.





{
    \small
    \bibliographystyle{ieeenat_fullname}
    \bibliography{main}
}
\end{document}